%% file: main.tex
\documentclass[dvipsnames,format=sigconf,anonymous=false,review=false]{acmart}
\usepackage{utils}

\AtBeginDocument{%
  \providecommand\BibTeX{{%
    \normalfont B\kern-0.5em{\scshape i\kern-0.25em b}\kern-0.8em\TeX}}}

\setcopyright{acmcopyright}
\copyrightyear{2018}
\acmYear{2018}
\acmDOI{10.1145/1122445.1122456}

\copyrightyear{2024}
\acmYear{2024}
\setcopyright{acmlicensed}\acmConference[GECCO '24]{Genetic and
Evolutionary Computation Conference}{July 14--18, 2024}{Melbourne, VIC,
Australia}
\acmBooktitle{Genetic and Evolutionary Computation Conference (GECCO '24),
July 14--18, 2024, Melbourne, VIC, Australia}
\acmDOI{10.1145/3638529.3654198}
\acmISBN{979-8-4007-0494-9/24/07}

\begin{document}

\title{CatCMA : Stochastic Optimization for Mixed-Category Problems}



\author{Ryoki Hamano}
\email{hamano_ryoki_xa@cyberagent.co.jp}
\orcid{0000-0002-4425-1683}
\affiliation{%
  \institution{CyberAgent, Inc. \and Yokohama National University}
  \city{Shibuya}
  \state{Tokyo}
  \country{Japan}
  \postcode{150-0042}
}

\author{Shota Saito}
\email{saito-shota-bt@ynu.jp}
\orcid{0000-0002-9863-6765}
\affiliation{%
  \institution{Yokohama National University \and SKILLUP NeXt Ltd.}
  \city{Yokohama}
  \state{Kanagawa}
  \country{Japan}
  \postcode{240-8501}
}

\author{Masahiro Nomura}
\email{nomura\_masahiro@cyberagent.co.jp}
\orcid{0000-0002-4945-5984}
\affiliation{%
  \institution{CyberAgent, Inc.}
  \city{Shibuya}
  \state{Tokyo}
  \country{Japan}
  \postcode{150-0042}
}

\author{Kento Uchida}
\email{uchida-kento-fz@ynu.ac.jp}
\orcid{0000-0002-4179-6020}
\affiliation{%
  \institution{Yokohama National University}
  \city{Yokohama}
  \state{Kanagawa}
  \country{Japan}
  \postcode{240-8501}
}

\author{Shinichi Shirakawa}
\email{shirakawa-shinichi-bg@ynu.ac.jp}
\orcid{0000-0002-4659-6108}
\affiliation{%
  \institution{Yokohama National University}
  \city{Yokohama}
  \state{Kanagawa}
  \country{Japan}
  \postcode{240-8501}
}

\renewcommand{\shortauthors}{Hamano and Saito, et al.}

\begin{abstract}
Black-box optimization problems often require simultaneously optimizing different types of variables, such as continuous, integer, and categorical variables. Unlike integer variables, categorical variables do not necessarily have a meaningful order, and the discretization approach of continuous variables does not work well. Although several Bayesian optimization methods can deal with mixed-category black-box optimization (MC-BBO), they suffer from a lack of scalability to high-dimensional problems and internal computational cost.
This paper proposes CatCMA, a stochastic optimization method for MC-BBO problems, which employs the joint probability distribution of multivariate Gaussian and categorical distributions as the search distribution. CatCMA updates the parameters of the joint probability distribution in the natural gradient direction. CatCMA also incorporates the acceleration techniques used in the covariance matrix adaptation evolution strategy (CMA-ES) and the stochastic natural gradient method, such as step-size adaptation and learning rate adaptation. In addition, we restrict the ranges of the categorical distribution parameters by margin to prevent premature convergence and analytically derive a promising margin setting. Numerical experiments show that the performance of CatCMA is superior and more robust to problem dimensions compared to state-of-the-art Bayesian optimization algorithms.


\end{abstract}



\maketitle


\input{body/01_intro} 
\input{body/02_igo} 
\input{body/03_proposed} 
\input{body/04_margin} 

\input{body/05_experiments} 
\input{body/06_conclusion} 


\begin{acks}
This work was partially supported by JSPS KAKENHI (JP23KJ0985, JP23H00491, JP23H03466), JST PRESTO (JPMJPR2133), and NEDO (JPNP18002, JPNP20006).
\end{acks}

\bibliographystyle{ACM-Reference-Format}
\bibliography{reference}

\clearpage 
\appendix
\input{body/99_appendix} 

\end{document}

%% file: body/01_intro.tex
\section{Introduction} \label{sec:intro}
Black-box optimization problems often require simultaneously optimizing different types of variables, such as continuous, integer, and categorical variables.
These problems with mixed-type variables appear in real-world applications, such as hyperparameter optimization in machine learning~\cite{hyperparam_tune:hutter2019, hazan2018hyperparameter}, hardware design~\cite{hardware:lu2020, hardware:touloupas2022}, and drug discovery~\cite{drugdiscovery:negoescu2011, drugdiscovery:zhou2017}.
Since there are often dependencies between different types of variables in such optimization problems, it is desirable to develop methods that allow simultaneous optimization of mixed-type variables.

The mixed-category black-box optimization~(MC-BBO) problem, the target of this study, is the problem of simultaneously optimizing continuous and categorical variables under a black-box setting.
Unlike integer variables, categorical variables do not necessarily have a meaningful order, and the discretization approach of continuous variables does not work well.
Therefore, mixed-category variables are more intractable than mixed-integer variables.

Although various black-box optimization methods have been proposed for the continuous and categorical domains, they are difficult to simply combine for MC-BBO because they are often developed separately for each domain and work based on different mechanisms\del{are quite different}{}.
The \emph{covariance matrix adaptation evolution strategy}~(CMA-ES)~\cite{hansen_adapting_1996, hansen_reducing_2003, hansen2016cma} has shown excellent performance in optimizing continuous variables. The CMA-ES is a stochastic continuous optimization method that utilizes the multivariate Gaussian distribution, and updates the mean vector, covariance matrix, and step-size (overall standard deviation) to minimize the objective function values. The update rules for the distribution parameters used in the CMA-ES are a combination of the weighted recombination of the mean vector, rank-$\mu$ and rank-one updates of the covariance matrix, and step-size adaptation.  Among them, the weighted recombination and rank-$\mu$ update are closely related to the stochastic natural gradient method~\cite{akimoto2012theoretical}, and can be derived from \emph{information geometric optimization}~(IGO)~\cite{IGO:2017} which is a unified framework of probabilistic model-based black-box algorithms based on the stochastic natural gradient method.
The \emph{adaptive stochastic natural gradient} (ASNG) method~\cite{ASNG:2019} is a stochastic optimization method that was originally developed for optimizing categorical variables for neural architecture search.\footnote{In \citet{ASNG:2019}, integer variables are optimized along with categorical variables using the Gaussian distribution with the diagonal covariance matrix under expectation parameterization. However, the parameterization of the distribution is different from that of CMA-ES and cannot include sophisticated update rules such as the rank-one update and step-size adaptation.}
The ASNG updates distribution parameters (the probability of generating each category\del{)}{} from a multivariate categorical distribution\nnew{) along the estimated natural gradient}\del{, assuming that categorical variables are independent in each dimension}{}. 
The ASNG update rule is also derived from the IGO framework and incorporates the learning rate adaptation based on the trust region.
Although the CMA-ES and the ASNG are related through the IGO framework, they were developed separately and have never been combined for MC-BBO.\del{, as mentioned above.}{}

In Bayesian optimization~\cite{frazier2018tutorial}, \emph{CASMOPOLITAN}~\cite{casmopolitan:2021} and \emph{TPE}~\cite{TPE:2011} are state-of-the-art methods for solving MC-BBO problems.
However, these methods face scalability issues and incur large internal computational costs.

In this \del{paper}{}\nnew{study}, we propose CatCMA, a stochastic optimization method for MC-BBO problems.
CatCMA employs the joint probability distribution of multivariate Gaussian and categorical distributions to generate mixed-category variables tractably.
The basic update rule of CatCMA is obtained by applying a joint probability distribution to the IGO framework.
To improve the optimization performance of CatCMA, we introduce rank-one update and step-size adaptation to the optimization of continuous variables.
In addition, to balance the optimization of continuous and categorical variables, CatCMA incorporates the learning rate adaptation employed in the ASNG.
Although the lower bound of distribution parameters, called \emph{margin}, is introduced into the categorical distribution to prevent premature convergence, we demonstrate that inappropriate margins degrade the performance of CatCMA.
To handle this problem, we analytically derived promising margin settings.
The numerical experiment using benchmark functions shows that the performance of CatCMA is superior and more robust to problem dimensions compared to CASMOPOLITAN and TPE.

%% file: body/02_igo.tex
\section{Information Geometric Optimization} \label{sec:IGO}
The information geometric optimization (IGO)~\cite{IGO:2017} is a unified framework for probabilistic model-based black-box algorithms.
Let $\{P_{\btheta}\}$ be a family of probability distributions on a search space~$\mathcal{X}$ parameterized by $\btheta \in \Theta$.
The IGO transforms the original problem into the maximization of the expected value $J_{\param[t]}$ of the utility function $W_{\param[t]}^f$.
The function $J_{\param[t]}$ is defined as
\begin{align}
    J_{\param[t]} (\btheta) = \int_{\x \in \mathcal{X}} W_{\param[t]}^f (\x) p(\x \mid \btheta) \diff \x \enspace, \label{eq:igo_argmax}
\end{align}
where $W_{\param[t]}^f (\x)$ is the utility function, and $p(\x \mid \btheta)$ is the probability density~(mass) function of $P_{\btheta}$.
The IGO maximizes \eqref{eq:igo_argmax} by natural gradient ascent. The natural gradient $\ngr_{\btheta} J_{\param[t]} (\btheta)$, which is defined as the product of the inverse of Fisher information matrix $F^{-1}(\btheta)$ and the vanilla gradient of $J_{\param[t]} (\btheta)$, is calculated as
\begin{align}
    \ngr_{\btheta} J_{\param[t]} (\btheta) = \int_{\x \in \mathcal{X}} W_{\param[t]}^f (\x) p(\x \mid \btheta) \ngr_{\btheta} \ln p(\x \mid \btheta)  \diff \x \enspace. \label{eq:IGO-ngr}
\end{align}
The IGO approximates the natural gradient $\ngr_{\btheta} J_{\param[t]} (\btheta)$ at $\btheta = \param[t]$ by the Monte Carlo estimation using $\lambda$ samples $\x_1, \ldots, \x_\lambda$ generated from $p(\x \mid \param[t])$ as
\begin{align}
    \left. \ngr_{\btheta} J_{\param[t]} (\btheta) \right|_{\btheta = \btheta^{(t)}} \approx \frac{1}{\lambda} \sum_{i=1}^\lambda w_i \left. \ngr_{\btheta} \ln p(\x_{i:\lambda} \mid \btheta) \right|_{\btheta = \param[t]} \enspace,
\end{align}
where the weight $w_i$ is the estimated utility function value \linebreak$W^f_{\param[t]}(\x_{i:\lambda})$.
The index of the $i$-th best sample is denoted as $i\!:\!\lambda$.
The update rule of the IGO is given as
\begin{align}
    \param[t+1] = \param[t] + \eta  \sum_{i=1}^\lambda \frac{w_i}{\lambda} \left. \ngr_{\btheta} \ln p(\x_{i:\lambda} \mid \btheta) \right|_{\btheta = {\param[t]}} \enspace,
\end{align}
where $\eta \in \R_{>0}$ is the learning rate.
The IGO framework covers several well-known probabilistic model-based optimization methods.
By applying the multivariate Gaussian distribution to the IGO framework, the pure rank-$\mu$ update CMA-ES~\cite{hansen_reducing_2003} can be obtained.
In addition \new{to the IGO update}, several methods, including CMA-ES and ASNG, employ acceleration techniques to realize practical optimization performance.

%% file: body/03_proposed.tex
\section{Proposed Method} \label{sec:proposed}
In this study, we propose a stochastic optimization method for mixed-category black-box optimization, named CatCMA.
Categorical variables do not necessarily have a meaningful order and are not suitable for being generated through discretization of continuous variables.
Therefore, to generate mixed-category variables tractably, we consider the joint probability distribution of multivariate Gaussian and categorical distributions.
The basic update rule of CatCMA is obtained by applying a joint probability distribution to the IGO framework.
This section defines the search space and the joint probability distribution and derives the basic update rule from the IGO framework.
In addition, we provide implementations for more practical optimization.
Algorithm~\ref{alg:proposed} shows the optimization process of CatCMA and Table~\ref{table:hyperparameter} shows the recommended hyperparameter settings.

\begin{algorithm}[t] 
    \caption{CatCMA}
    \begin{algorithmic}[1] \label{alg:proposed}
        \REQUIRE $\m[0], \sig[0], \C[0], \qt[0]$
        \STATE $\ps[0] = \pc[0] = \boldsymbol{0}$, $\delt[0] = 1$, $\st[0] = \boldsymbol{0}$, $\gamt[0] = 0$, $t=0$
        \WHILE{termination conditions are not met}
        \FOR{$i=1,\ldots,\lambda$}
            \STATE Sample $\x_i \sim p(\x \mid \m[t], (\sig[t])^2\C[t])$
            \STATE Sample $\bc_i \sim p(\bc \mid \qt[t])$
        \ENDFOR
        \STATE Evaluate $(\x_1, \bc_1), \ldots, (\x_\lambda, \bc_\lambda)$ on $f(\x, \bc)$.
        \STATE Update $\m[t]$ using \eqref{eq:igo_m}.
        \STATE Update $\ps[t]$ and $\pc[t]$ using \eqref{eq:update_ps} and \eqref{eq:update_pc}.
        \STATE Update $\C[t]$ and $\sig[t]$ using \eqref{eq:update_C} and \eqref{eq:update_sig}.
        \STATE Update $\qt[t]$ using \eqref{eq:update_q}
        \STATE Update $\st[t]$ and $\gamt[t]$ using \eqref{eq:update_s} and \eqref{eq:update_gam}.
        \STATE Update $\delt[t]$ using \eqref{eq:update_del}.
        \STATE Modify $\sig[t+1]$ using \eqref{eq:post-process_sigma}.
        \STATE Modify $\qt[t+1]$ using \eqref{eq:margin_correction_1} and \eqref{eq:margin_correction_2}.
        \STATE $t \leftarrow t+1$
        \ENDWHILE
    \end{algorithmic} 
\end{algorithm}

\renewcommand{\arraystretch}{2.1}
\begin{table}[t]
\centering
\caption{The hyperparameter setting of CatCMA.}
\begin{tabular}{l}
\hline
$\lambda = 4 + \lfloor 3 \ln (\Nco + \Nca) \rfloor$, \enspace $\mu = \lfloor \lambda/2 \rfloor$ \\
$\frac{w_i}{\lambda} = \frac{\ln (\frac{\lambda+1}{2}) - \ln i}{ \sum_{j=1}^{\mu} \left(\ln (\frac{\lambda+1}{2}) - \ln j\right)}$ if $i \leq \mu$, otherwise $0$. \\
$\muw = \left(\sum_{i=1}^\mu \left(\frac{w_i}{\lambda}\right)^2 \right)^{-1}$, \enspace $c_m = 1$, \enspace $c_\sigma = \frac{\muw + 2}{\Nco + \muw + 5}$ \\
$d_\sigma = 1 + c_\sigma + 2 \max \left(0, \sqrt{\frac{\muw - 1}{\Nco +1}} - 1 \right)$, \enspace $c_c = \frac{4 + \muw/{\Nco}}{\Nco + 4 + 2\muw/{\Nco}}$\\
$c_1 = \frac{2}{(\Nca + 1.3)^2 + \muw}$, \enspace $c_\mu = \min\left( 1-c_1, \frac{2(\muw - 2 + 1/\muw)}{(\Nco+2)^2 + \muw} \right)$ \\
$\alpha = 1.5$, \enspace $\Lambda^{\min} = 10^{-30}$, \enspace $\qmin_n = (1-0.73^{\frac{1}{\Nca}})/(K_n - 1)$ \\
\hline
\end{tabular}
\label{table:hyperparameter}
\end{table}
\renewcommand{\arraystretch}{1.0}

\subsection{Search Space and Joint Probability Distribution}
\paragraph{Definition of Search Space}
We consider the mixed-category objective function $f$ whose first $\Nco$ variables are continuous and the rest $\Nca$ variables are categorical.
The search space of $f$ is given by $\mathcal{X} \times \mathcal{C}$, where $\mathcal{X} = \R^{\Nco}$ and $\mathcal{C} = \mathcal{C}_1 \times \cdots \times \mathcal{C}_{\Nca}$. The $n$-th categorical domain $\mathcal{C}_n$ is defined as
\begin{align}
    \mathcal{C}_n = \left\{ \bc_n \in \{0, 1\}^{K_n} \enspace \middle| \enspace \sum_{k=1}^{K_n} \bc_{n, k} = 1 \right\} \enspace,
\end{align}
where $K_n$ is the number of categories in the $n$-th categorical variable, and $\bc_n$ is a $K_n$-dimensional one-hot vector.
We denote that $\bc_{n,k}$ means the $k$-th element of $\bc_{n}$.

\paragraph{Joint Probability Distribution for CatCMA}
We define a joint probability distribution of the multivariate Gaussian distribution and categorical distribution on the search space $\mathcal{X} \times \mathcal{C}$.
First, the multivariate Gaussian distribution on $\mathcal{X}$ is given as
\begin{align*}
    p (\x \mid \boldsymbol{m}, \boldsymbol{C} ) = \frac{1}{(2 \pi)^{\Nco/2} \det (\boldsymbol{C})^{1/2}} \exp \left( - \frac12 (\x - \boldsymbol{m})^\top \boldsymbol{C}^{-1} (\x - \boldsymbol{m}) \right) ,
\end{align*}
where $\boldsymbol{m} \in \R^{\Nco}$ is the mean vector, and $\boldsymbol{C} \in \R^{\Nco \times \Nco}$ is the covariance matrix.
Next, the family of categorical distributions on $\mathcal{C}$ is given as
\begin{align*}
    p (\bc \mid \q) = \prod_{n=1}^{\Nca} \prod_{k=1}^{K_n} (\q_{n,k})^{\bc_{n,k}} \enspace.
\end{align*}
The distribution parameter $\q$ is defined as \nnew{$\q = (\q_1^\top, \ldots, \q_{\Nca}^\top)^\top$ with}
\begin{align}
    \q_n \in [0, 1]^{K_n} \enspace \text{s.t.} \enspace \sum_{k=1}^{K_n} \q_{n,k} = 1 \enspace,
\end{align}
where $\q_{n,k}$ represents the probability of $\bc_{n, k} = 1$.
Finally, the joint probability distribution is given as follows:
\begin{align}
    p (\x, \bc \mid \boldsymbol{m}, \boldsymbol{C}, \q) = p(\x \mid \boldsymbol{m}, \boldsymbol{C}) p (\bc \mid \q)
\end{align}

\subsection{Derivation of IGO Update with Joint Probability Distribution}
We derive the update rule of the distribution parameter by applying the joint probability distribution to the IGO framework. First, we redefine the $n$-th categorical distribution parameter as follows:\footnote{If we define $\q_n = (\q_{n, 1}, \ldots, \q_{n, K_n})$, the Fisher information matrix $F(\btheta)$ is degenerate because of the linear dependence $\sum_{k=1}^{K_n} \q_{n,k} = 1$.}
\begin{align}
    \q_n = ( \q_{n, 1}, \ldots, \q_{n, K_n - 1} )^\top
\end{align}
We note that the probability of $\bc_{n, K_n} = 1$ is determined as $\q_{n, K_n} = 1 - \sum_{k=1}^{K_n - 1} \q_{n,k}$.
The distribution parameter of the joint probability distribution $\btheta$ is defined as
\begin{align}
    \btheta = \left(\boldsymbol{m}^\top, \VEC(\boldsymbol{C})^\top, \q_1^\top, \ldots, \q_{\Nca}^\top \right)^\top \enspace,
\end{align}
where $\VEC(\cdot)$ rearranges a matrix into a column vector.
The Fisher information matrix of the joint probability distribution is given by the block diagonal matrix with $N_\mathrm{ca} + 2$ blocks as
\begin{align*}
    F(\btheta) = \diag \left(F(\boldsymbol{m}), F(\VEC(\boldsymbol{C})), F(\q_1), \ldots, F(\q_{\Nca}) \right) \enspace,
\end{align*}
where $F(\boldsymbol{m}) = \boldsymbol{C}^{-1}$, $F(\VEC(\boldsymbol{C})) = \frac12 \boldsymbol{C}^{-1}  \otimes \boldsymbol{C}^{-1}$, and $F(\q_{n}) = \diag(\q_{n})^{-1} + \left(1 - \sum_{k=1}^{K_{n} - 1} \q_{n,k} \right)^{-1} \boldsymbol{1} \boldsymbol{1}^\top$. We denote $\otimes$ as the Kronecker product. Moreover, the inverse of the Fisher information matrix can be calculated as
\begin{align*}
    F^{-1}(\btheta) = \diag \left(F^{-1}(\boldsymbol{m}), F^{-1}(\VEC(\boldsymbol{C})), F^{-1}(\q_1), \ldots, F^{-1}(\q_{\Nca}) \right) \enspace, 
\end{align*}
where $F^{-1}(\boldsymbol{m}) = \boldsymbol{C}$, $F^{-1}(\VEC(\boldsymbol{C})) = 2 \boldsymbol{C} \otimes \boldsymbol{C}$, and $F^{-1}(\q_n) = \diag (\q_{n}) + \q_{n} \q_{n}^\top$.
For the part corresponding to $\boldsymbol{m}$ and $\boldsymbol{C}$, the detailed derivation of $F(\btheta)$ and $F^{-1}(\btheta)$ can be found in \cite{Akimoto:2010}, and for the part corresponding to $\q_n$, derivation is in \cite{ASNG:2019}.
The vanilla gradients of the log-likelihood are calculated as follows.
\begin{align}
    \nabla_{\boldsymbol{m}} \ln p (\x, \bc \!\mid\! \btheta) &= \boldsymbol{C}^{-1}( \x - \boldsymbol{m} ) \\
    \nabla_{\! \VEC(\boldsymbol{C})} \! \ln p (\x, \bc \!\mid\! \btheta) &= \frac12 \VEC \! \left ( \boldsymbol{C}^{-1} \! (\x - \boldsymbol{m}) (\x - \boldsymbol{m})^{\!\top} \!\boldsymbol{C}^{-1}\!\! -\! \boldsymbol{C}^{-1} \right) \!\\
    \nabla_{\q_{n,k}} \ln p (\x, \bc \!\mid\! \btheta) &= \frac{\bc_{n,k}}{\q_{n,k}} - \frac{\bc_{n, K_n}}{1 - \sum_{k'=1}^{K_n - 1} \q_{n,k'}}
\end{align}
The natural gradient of the log-likelihood is given by the product of the inverse of the Fisher information matrix and the vanilla gradient of the log-likelihood.
Finally, introducing learning rates $c_m$ and $c_\mu$, as well as $\eta_1, \ldots, \eta_{\Nca}$ for each block, the update rule of the distribution parameter is given as follows.
\begin{align}
    \m[t+1] \!\! &= \m[t] + c_m \sum_{i=1}^{\lambda} \frac{w_{i}}{\lambda} ( \x_{i:\lambda} - \m[t] ) \label{eq:igo_m} \\
    \C[t+1] \!\! &= \C[t] \! + \! c_\mu \! \sum_{i=1}^{\lambda} \! \frac{w_{i}}{\lambda} \! \Bigl( (\x_{i:\lambda} - \m[t]) (\x_{i:\lambda} - \m[t])^{\! \top} \!\! - \C[t] \Bigr) \label{eq:igo_C} \\
    \qt[t+1]_n \!\! &= \qt[t]_n + \eta_n \sum_{i=1}^\lambda \frac{w_i}{\lambda} (\bc_{i:\lambda, n} - \qt[t]_n) \label{eq:igo_q}
\end{align}

\subsection{Introducing Enhancement Mechanisms} \label{ssec:adaptation}

\new{
Several probabilistic model-based optimization methods employ additional components to the IGO update to achieve practical optimization performance.
We note that the derived update rules \eqref{eq:igo_m} and \eqref{eq:igo_C} are equivalent to the mean vector update and rank-$\mu$ update of the CMA-ES~\cite{hansen_adapting_1996, hansen2016cma}.
The CMA-ES also contains the rank-one update and step-size adaptation, which allow for faster adaptation of the multivariate Gaussian distribution.
To improve the optimization performance, we introduce these mechanisms to CatCMA.
}

\new{
In addition to the performance improvement, CatCMA has another issue in the sensitivity to the hyperparameter setting.
The performance of the IGO algorithm is strongly influenced by the hyperparameter setting, and this trend is more pronounced for the optimization of the joint probability distribution.
With an inappropriate hyperparameter setting, the updates of the multivariate Gaussian and categorical distributions become unbalanced, and the optimization will fail.
Because suitable learning rates vary with the objective function and algorithm state, the comprehensive investigation for suitable settings is not promising.
To address this problem, CatCMA incorporates the learning rate adaptation employed in the ASNG, which is introduced into the update in \eqref{eq:igo_q}.
}

\subsubsection{Rank-one update and step-size adaptation for the covariance matrix update}

First, we introduce the step-size $\sig[t] \in \R_{>0}$, and the candidate solutions are sampled from the multivariate Gaussian distribution $p(\x \mid \m[t], (\sig[t])^2 \C[t])$ as
\begin{align}
    \y_i &\sim \mathcal{N}(\boldsymbol{0}, \C[t]) \enspace, \\
    \x_i &= \m[t] + \sig[t] \y_i \enspace.
\end{align}
We introduce the two evolution paths to cumulate the update direction of the distribution parameter.
\begin{align}
    \ps[t+1] &= (1-c_\sigma)\ps[t] + \sqrt{c_\sigma(2-c_\sigma)\muw} {\C[t]}^{-\frac12} \sum_{i=1}^\lambda \frac{w_i}{\lambda} \y_{i:\lambda} \enspace, \label{eq:update_ps} \\
    \pc[t+1] &= (1-c_c) \pc[t] + h_\sigma \sqrt{c_c(2-c_c)\muw} \sum_{i=1}^\lambda \frac{w_i}{\lambda} \y_{i:\lambda} \enspace, \label{eq:update_pc}
\end{align}
where $\muw$ denotes $1 / \sum_{i=1}^{\mu} w_i^2$; $c_\sigma$ and $c_c$ are cumulative rates; and $h_\sigma$ is $1$ if the following condition is satisfied, and $0$ otherwise.
\begin{align*}
    \|\ps[t+1]\| < \sqrt{1-(1-c_\sigma)^{2(t+1)}}\left(1.4+\frac{2}{\Nco+1}\right)\E [\|\mathcal{N}(\boldsymbol{0}, \boldsymbol{I}) \|]
\end{align*}
The expected norm $\E [\|\mathcal{N}(\boldsymbol{0}, \boldsymbol{I}) \|]$ is approximated by $\sqrt{\Nco} \bigl( 1 - \frac{1}{4 \Nco} + \frac{1}{21 N_\mathrm{co}^2} \bigr)$.
The covariance matrix $\C[t]$ is updated as
\begin{multline}
     \C[t+1] = \left(1 + (1-h_\sigma)c_1 c_c(2-c_c) \right) \C[t] \\
    + c_1 \left(\pc[t+1]{\pc[t+1]}^\top - \C[t] \right) + c_\mu \sum_{i=1}^{\lambda} \frac{w_i}{\lambda} \left( \y_{i:\lambda}\y_{i:\lambda}^\top - \C[t] \right) , \label{eq:update_C}
\end{multline}
where $c_1$ and $c_{\mu}$ are the learning rates for the rank-one update and the rank-$\mu$ update, respectively, which is consistent with \eqref{eq:igo_C}.
The step-size $\sig[t]$ is updated as
\begin{align}
    \sig[t+1] = \sig[t] \exp \left( \frac{c_\sigma}{d_\sigma} \left( \frac{\|\ps[t+1] \|}{\E [\|\mathcal{N}(\boldsymbol{0}, \boldsymbol{I}) \|]} - 1 \right) \right) \enspace, \label{eq:update_sig}
\end{align}
where $d_\sigma$ is a damping factor.
The hyperparameters used in the above steps are based on \cite[Table~2]{borenstein_principled_2014} and are summarized in Table~\ref{table:hyperparameter}.

\subsubsection{Learning rate adaptation for the categorical distribution update}
We introduce learning rate adaptation to keep the signal-to-noise ratio (SNR) of the approximated natural gradient in the ASNG somewhat large.
The rationale for using SNR in learning rate adaptation is discussed in \cite{nomura2024cma}.
Instead of the categorical distribution update in \eqref{eq:igo_q}, we employ the natural gradient ascent with gradient normalization,
\begin{align}
    \qt[t+1] = \qt[t] + \delt[t] \frac{G(\qt[t])}{\left\| G(\qt[t]) \right\|_{F(\qt)}} \enspace, \label{eq:update_q}
\end{align}
where $\delt[t]$ is the trust region radius; $\| \cdot \|_{F(\qt)}$ is the Fisher norm relative to $\qt$; and $G(\qt[t])$ is the estimated natural gradient given as
\begin{align}
    G(\qt[t]) = \sum_{i=1}^\lambda \frac{w_i}{\lambda} (\bc_{i:\lambda} - \qt[t]) \enspace.
\end{align}
We denoted $F(\qt[t]) = \diag\bigl( F(\qt[t]_1), \ldots, F(\qt[t]_{\Nca}) \bigr)$.
To adapt the trust region $\delt[t]$, we introduce the accumulation of the estimated natural gradient\footnote{The ASNG uses an approximation to compute $F^{\frac12}(\qt[t])$ because the computation of $F^{\frac12}(\qt[t])$ requires $O(\sum_{n=1}^{\Nca} (K_n - 1)^3)$ time complexity. The details of the calculation procedure can be found in \cite[Appendix B]{ASNG:2019}.} as
\begin{align}
    \st[t+1] &= (1 - \beta)\st[t] + \sqrt{\beta (2 - \beta)} F^{\frac12}(\qt[t]) G(\qt[t]) \enspace, \label{eq:update_s} \\
    \gamt[t+1] &= (1 - \beta)^2 \gamt[t] + \beta(2 - \beta) \| G(\qt[t]) \|_{F(\qt)}^2 \enspace, \label{eq:update_gam}
\end{align}
where $\beta$ is set to $\delt[t]/\bigl(\sum_{n=1}^{\Nca}(K_n - 1) \bigr)^{\frac12}$.
Using a constant $\alpha = 1.5$, the adaptation of $\delt[t]$ is done as follows:
\begin{align}
    \delt[t+1] = \delt[t] \exp \left( \beta \left( \frac{\| \st[t+1] \|^2}{\alpha} - \gamt[t+1] \right) \right) \label{eq:update_del}
\end{align}

\subsection{Post-processing for the Multivariate Gaussian Distribution}
When the multivariate Gaussian distribution converges earlier than the categorical distribution, the eigenvalues of the covariance matrix $(\sig[t])^2 \C[t]$ can be very small, which causes numerical errors. To prevent numerical errors, we introduce a post-process for the step-size $\sig[t+1]$ so that the eigenvalues do not become too small as
\begin{align}
    \sig[t+1] \leftarrow \max \left\{ \sig[t+1], ~ \sqrt{\frac{\Lambda^{\min}}{ \min\{\mathrm{eig}(\C[t+1])\} }} \right\} \enspace, \label{eq:post-process_sigma}
\end{align}
where $\mathrm{eig}(\C[t+1])$ is the set of the eigenvalues of $\C[t+1]$, and $\Lambda^{\min}$ is the lower bound of the eigenvalues of the multivariate Gaussian distribution. In this study, we set $\Lambda^{\min} = 10^{-30}$.

\subsection{Margin Correction for the Categorical Distribution}
\begin{figure*}[t]
    \centering
    \includegraphics[width=0.99\linewidth]{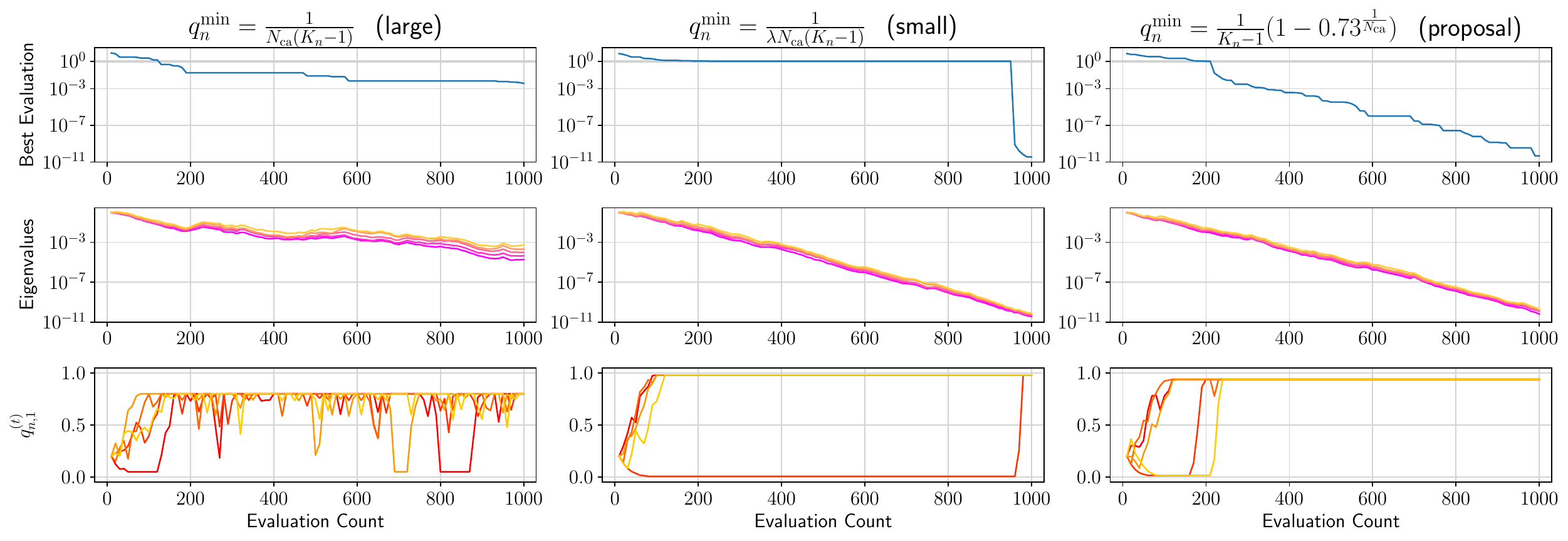}
    \caption{Transition of the best evaluation value, the eigenvalues of $(\sig[t])^2 \C[t]$, and probability of generating the best category $\qt[t]_{n,1}$ in one typical trial of optimizing \textsc{SphereCOM} with $\Nco = \Nca = 5$ and $K_n = 5$.}
    \label{fig:trans_margin}
\end{figure*}
If some elements of the categorical distribution parameter become too small, the corresponding categories are hardly generated, causing a stagnation of the optimization. Thus, CatCMA introduces the margin correction, which guarantees the lower bound of the distribution parameter. In this study, we employ the update procedure introduced in the ASNG as the margin correction.
\begin{align}
    \qt[t+1]_{n,k} &\leftarrow \max \{ \qt[t+1]_{n,k}, \qmin_n \} \label{eq:margin_correction_1} \\
    \qt[t+1]_{n,k} &\leftarrow \qt[t+1]_{n,k} + \frac{1 - \sum_{k'=1}^{K_n} \qt[t+1]_{n,k'} }{\sum_{k'=1}^{K_n} (\qt[t+1]_{n,k'} - \qmin_n)} (\qt[t+1]_{n,k} - \qmin_n) \label{eq:margin_correction_2}
\end{align}
The margin correction guarantees $\qt[t+1]_{n,k} \geq \qmin_n$ while keeping $\sum_{k'=1}^{K_n} \qt[t+1]_{n,k'} = 1$.
The margin $\qmin_n$ to be set is discussed in Section~\ref{sec:margin}.

%% file: body/04_margin.tex
\section{Promising Margin Setting} \label{sec:margin}
To prevent certain categories from not being generated, CatCMA introduces a margin to the categorical distribution parameter.
While the lower bound $\qmin_n$ is set to $1/(\Nca (K_n -1))$ in the ASNG, this setting is not necessarily appropriate for CatCMA as well.
We first demonstrate that optimization performance degrades if the margin is not set appropriately in CatCMA.
Through this observation, we derive a promising margin setting that stabilizes the optimization.

\subsection{Behaviours under Inappropriate Margins} \label{ssec:inappropriate}
In this section, we perform CatCMA under too large and too small margins and discuss the importance of margin setting in CatCMA through the results of preliminary experiments.
As a benchmark function, we use \textsc{SphereCOM}, which is defined in Table~\ref{table:benchmark}. The numbers of dimensions and categories were set as $\Nco = \Nca = 5$ and $K_1 = \ldots = K_{\Nca} = 5$.
The initial mean vector $\m[0]$ was set to $(1, \ldots, 1)$.
The initial step-size $\sig[0]$ was given by $1$, and the initial covariance matrix $\C[0]$ was given by an identity matrix.
The initial categorical distribution parameter was set as $\qt[0]_{n,k} = 1/K_n$.
The margin settings were then varied as $\qmin_n = 1/(\Nca (K_n -1))$ (large setting), $1/(\lambda \Nca (K_n -1))$ (small setting), and $(1-0.73^{\frac{1}{\Nca}})/(K_n - 1)$ (recommended setting derived in the next subsection).

Figure~\ref{fig:trans_margin} shows the transition of the best evaluation value, eigenvalues of $(\sig[t])^2 \C[t]$, and probability of generating the best category $\qt[t]_{n,1}$.
Focusing on the result for large setting $\qmin_n = 1/(\Nca \linebreak (K_n -1))$, which is the default value used in the ASNG, we observe that until the 200 evaluation count, CatCMA finds a solution where all categorical variables are optimal.
This can be seen from the fact that $\Nca - \sum_{n=1}^{\Nca} \bc_{n,1}$ in \textsc{SphereCOM} takes only non-negative integers and the best evaluation value is less than $1$.
However, the convergence of the multivariate Gaussian distribution is stalled, resulting in poor improvement of the best evaluation value.
The probabilities of generating the optimal categories cannot be sufficiently high because of the large margin. Therefore, the candidate solutions used to update the distribution parameter include non-optimal categories.
Because selecting non-optimal categories leads to a lower ranking of candidate solution even if it contains desirable continuous variables, overly large margin setting has a negative impact on the update of the multivariate Gaussian distribution.

The results of small margin $\qmin_n = 1/(\lambda \Nca (K_n -1))$ indicate that the categorical distribution parameters are partly fixed, hardly generating optimal categories.
For the distribution parameters to be updated in the desired direction, the optimal categories need to be generated.
Thus, after the generated probabilities for optimal categories become small under an overly small margin, the fixation of the categorical distribution parameter is difficult to resolve.

These observations lead to the requirement of an appropriate margin value that would not hamper the convergence of the multivariate Gaussian distribution, preventing the categorical distribution from being fixed.

\begin{table*}[t]
    \caption{Benchmark functions}
    \label{table:benchmark}
    \centering
    \renewcommand{\arraystretch}{1.8}
    \begin{tabular}{cl}
        \hline
        Name & Definition \\
        \hline \hline
        \textsc{SphereCOM} & $f(\x, \bc) = \sum_{n=1}^{\Nco} \x_n^2 + \Nca - \sum_{n=1}^{\Nca} \bc_{n,1}$ \\
        \textsc{RosenbrockCLO} & $f(\x, \bc) = \sum_{n=1}^{\Nco -1} \Bigl(100 \bigl( \x_n^2 - \x_{n+1} \bigr)^2 + \bigl( \x_n - 1 \bigr)^2  \Bigr) + \Nca - \sum_{n=1}^{\Nca} \prod_{n' = 1}^n \bc_{n', 1}$ \\
        \textsc{MCProximity} & $f(\x, \bc) = \sum_{n=1}^{\Nco (= \Nca)} \bigl( \x_n - \z_n \bigr)^2 + \sum_{n=1}^{\Nca} \z_n$ \enspace, \enspace where $\z_n = \frac{1}{K_n}\sum_{k=1}^{K_n} (k-1) \bc_{n,k}$ \enspace. \\
        \hline
    \end{tabular}
    \renewcommand{\arraystretch}{1.0}
\end{table*}

\subsection{Derivation of Promising Margin}
We observed that when the margin is large, the probability of generating samples containing non-optimal categories increases, which prevents the convergence of the multivariate Gaussian distribution.
To address this issue, we first focus on the fact that CatCMA uses only the best $\lfloor \lambda/2 \rfloor$ samples to update the distribution parameters since the weight $w_i$ is set to $0$ when $i > \lfloor \lambda/2 \rfloor$.
To discuss the condition for promising margin setting, we consider the case where only $\lambda_\mathrm{non}$ samples contain the non-optimal categories, and the remaining $\lambda_\mathrm{opt} := \lambda - \lambda_\mathrm{non}$ samples have better evaluation values than any of the $\lambda_\mathrm{non}$ samples.
We also assume that the non-optimal categories determine the gap between the $\lambda_\mathrm{non}$ and $\lambda_\mathrm{opt}$ samples.
In this case, because the multivariate Gaussian and categorical distributions are independent, the update of the multivariate Gaussian distribution in CatCMA corresponds to the update of CMA-ES using $\lambda_\mathrm{opt}$ samples with the positive weights for best $\mu = \lfloor \lambda/2 \rfloor$ samples when $\lambda_\mathrm{opt} \geq \lfloor \lambda/2 \rfloor$, that is, $\lambda_\mathrm{non} \leq \lambda - \lfloor \lambda/2 \rfloor$. 
Consequently, in the following proposition, we derive a margin to guarantee that the number of samples containing non-optimal categories is less than $\lambda - \lfloor \lambda/2 \rfloor$ with high probability.
\begin{proposition} \label{prop:promising_margin}
Without loss of generality, assuming that categories of the optimal solution are the first categories in all dimensions and the parameter of the categorical distribution satisfies
\begin{align}
    \begin{split}
        &\qt[t]_{n,1} = 1 - \qmin_n (K_n - 1) \enspace, \\
        &\qt[t]_{n, k} = \qmin_n \quad \text{for all} \enspace k \in \{2, \ldots, K_n \}
    \end{split} \label{eq:assum_cat}
\end{align}
for all $n \in \{1, \ldots, \Nca\}$.
Let $\lambda_\mathrm{non}$ be the random variable that counts the number of samples containing non-optimal categories among the $\lambda$ samples.
If for a constant $\xi = 0.27$, the margin satisfies
\begin{align}
    \qmin_n = \frac{1}{K_n - 1} \left( 1 - (1-\xi)^{\frac{1}{\Nca}} \right) \label{eq:assum_margin}
\end{align}
for all $n \in \{1, \ldots, \Nca\}$, it holds, for any $\lambda \geq 6$,
\begin{align}
    \Pr\left(\lambda_\mathrm{non} \leq \lambda - \left\lfloor \frac{\lambda}{2} \right\rfloor \right) \geq 0.95 \enspace.
\end{align}
\end{proposition}
The proof of Proposition~\ref{prop:promising_margin} can be found in the supplementary material.
Proposition~\ref{prop:promising_margin} provides the margin such that the candidate solutions used to update the distribution parameter do not contain the non-optimal categories with high probability after the categorical distribution converges.
Hence, if the margin is large enough not to exceed $(1 - 0.73^{\frac{1}{\Nca}})/(K_n - 1)$, it prevents the categorical distribution from becoming fixed without hampering the convergence of the multivariate Gaussian distribution.
Figure~\ref{fig:trans_margin} shows that when the margin is set according to \eqref{eq:assum_margin}, the fixation and stagnation issues are resolved.
The detailed evaluation of the derived margin is shown in Section~\ref{ssec:exp_margin}.

%% file: body/05_experiments.tex
\section{Experiments} \label{sec:experiments}
We evaluated the search performance of CatCMA on the mixed-category problems. The code for CatCMA will be made available at \textcolor{blue}{\url{https://github.com/CyberAgentAILab/cmaes}}~\cite{nomura2024cmaes}.

\subsection{Benchmark Functions}
In this experiment, we used the following benchmark functions.
\begin{itemize}
    \item SphereCategoricalOneMax (\textsc{SphereCOM})
    \item RosenbrockCategoricalLeadingOnes (\textsc{RosenbrockCLO})
    \item MixedCategoryProximity (\textsc{MCProximity})
\end{itemize}
These definitions are summarised in Table~\ref{table:benchmark}. The two functions \text{SphereCOM} and \textsc{RosenbrockCLO} are the sum of benchmark functions for continuous and categorical optimization. On the other hand, \textsc{MCProximity} has a dependency between continuous and categorical variables.

\begin{figure*}[t]
    \centering
    \includegraphics[width=0.98\linewidth]{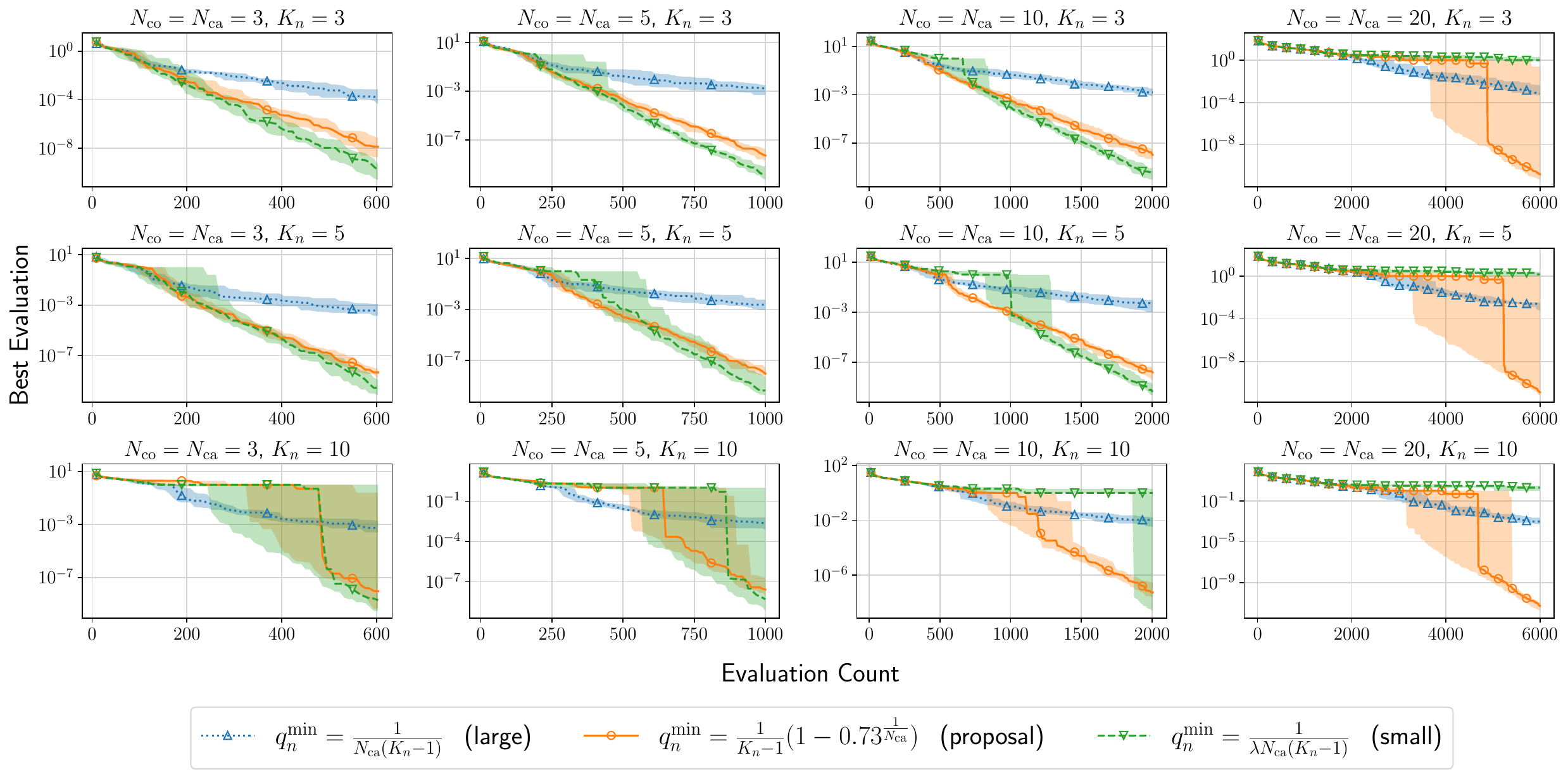}
    \caption{Transition of the best evaluation value on \textsc{SphereCOM}. The line and shaded area denote the medians and interquartile ranges over 20 independent trials, respectively.}
    \label{fig:spherecom_grid}
\end{figure*}

\subsection{Effectiveness of the Derived Margin} \label{ssec:exp_margin}
We derived the promising margin setting to stabilize the optimization in Section~\ref{sec:margin}. Firstly, we evaluated the effectiveness of the derived margins by varying the number of dimensions and categories.

\paragraph{Experimental Setting}
We used \textsc{SphereCOM} as the benchmark function.
The number of dimensions for continuous and categorical variables, $\Nco$ and $\Nca$, were set to $3$, $5$, $10$, and $20$.
The number of categories $K_n$ was equal in all dimensions and was set to $3$, $5$ and $10$.
The initial mean vector $\m[0]$ was uniformly sampled in $[-3, 3]^{\Nco}$.
The initial step-size $\sig[0]$ was set to $1$ and the initial covariance matrix $\C[0]$ was set to an identity matrix.
The initial categorical distribution parameter was set as $\qt[0]_n = 1/K_n$.
As margins, we used $1/(\Nca (K_n - 1))$ (large setting), $1/(\Nca (\lambda K_n - 1))$ (small setting), and $(1 - 0.73^{1/\Nca})/(K_n - 1)$ (recommended setting).
We performed 20 independent trials for each setting.

\paragraph{Results and Discussion}
Figure~\ref{fig:spherecom_grid} shows the transition of the best evaluation value on \textsc{SphereCOM}. 
When the margin is the large setting, the best evaluation value is stalled, as discussed in Section~\ref{ssec:inappropriate}.
The small margin setting shows good performance when $\Nco$, $\Nca$ are small, but suffers from the stagnation when $\Nco$, $\Nca$ are large.
On the other hand, when the margin is the recommended setting, the evaluation value steadily improve although it might be slightly degraded to the small setting.
In particular, the derived margin is better as $\Nco$, $\Nca$, and $K_n$ become large.
Therefore, for the robustness to the number of dimensions and categories, $\qmin_n = (1 - 0.73^{1/\Nca})/(K_n - 1)$ is recommended as a default parameter and is used in the subsequent experiment.
The results on \textsc{RosenbrockCLO} and \textsc{MCProximity} can be found in the supplementary material.

\subsection{Comparison with Baseline Methods}
To evaluate the effectiveness of CatCMA, we performed the comparative experiment with the baseline methods for mixed-category optimization.

\paragraph{Experimental Setting}

We varied the number of dimensions for continuous and categorical variables and the number of categories as $(\Nco, \Nca, K_n) = (3, 3, 3), (5, 5, 5), (10, 10, 10)$.
To confirm the effectiveness of CatCMA in mixed-category problems, we experimented with CASMOPOLITAN~\cite{casmopolitan:2021}, the state-of-the-art method in this problem setting, and Tree-structured Parzen Estimator~(TPE)~\cite{TPE:2011} implemented by optuna~\cite{optuna:akiba2019}. 
In the two methods, the range of the continuous variables was set to $[-3, 3]$, and hyperparameters were set to default ones.
The evaluation budget on the CASMOPOLITAN was set to 400 due to long internal computation time.
In addition, to confirm the effectiveness of the enhancement mechanisms in CatCMA, we experimented with CatCMA without enhancement mechanisms introduced in Section~\ref{ssec:adaptation}. In CatCMA without enhancement mechanisms, the modification of \eqref{eq:post-process_sigma} is not used and the learning rates are set to $c_m = 1$, $c_\mu = \min\left( 1-c_1, \frac{2(\muw - 2 + 1/\muw)}{(\Nco+2)^2 + \muw} \right)$, and $\eta_n = 1/(\Nca (K_n -1))$.
In CatCMA with and without enhancement mechanisms, the initial distribution parameters were set as in Section~\ref{ssec:exp_margin}, and the margin was set as $\qmin_n = (1-0.73^{\frac{1}{\Nca}})/(K_n -1)$.
We performed 20 independent trials for each setting.

\begin{figure*}[t]
    \centering
    \includegraphics[width=0.95\linewidth]{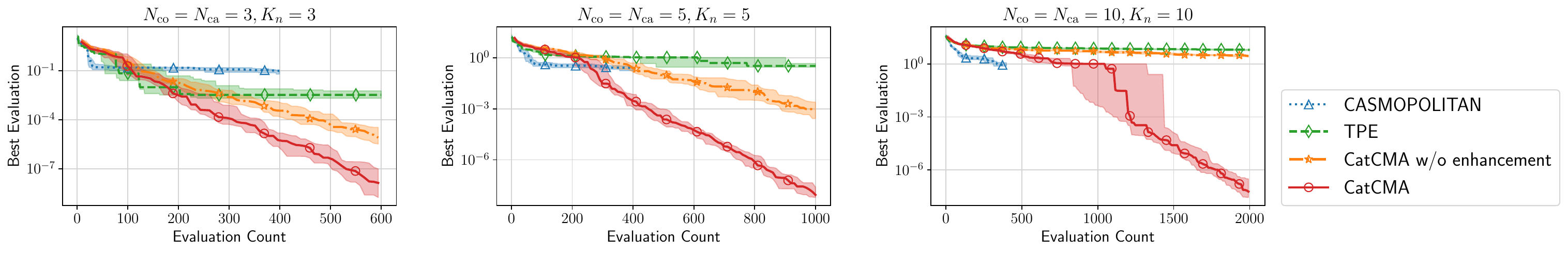}
    \caption{Transition of the best evaluation value on \textsc{SphereCOM}. The line and shaded area denote the medians and interquartile ranges over 20 independent trials, respectively.}
    \label{fig:spherecom_all}
\end{figure*}

\begin{figure*}[t]
    \centering
    \includegraphics[width=0.95\linewidth]{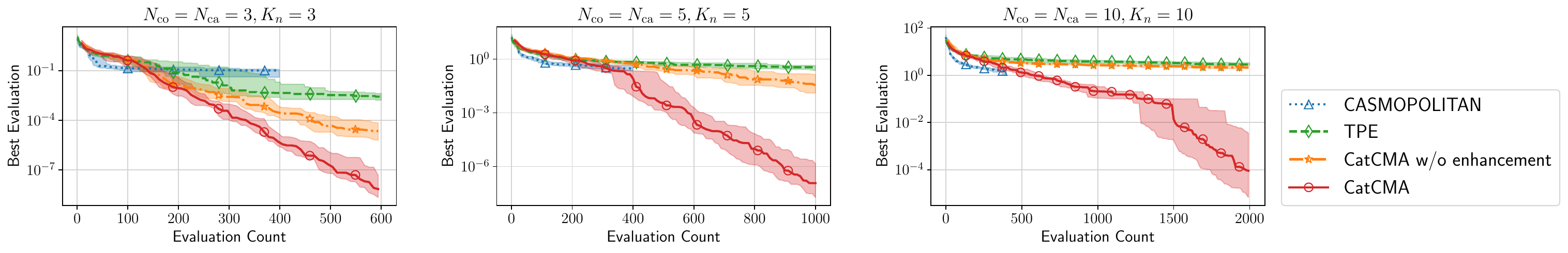}
    \caption{Transition of the best evaluation value on \textsc{MCProximity}. The line and shaded area denote the medians and interquartile ranges over 20 independent trials, respectively.}
    \label{fig:mcproximity_all}
\end{figure*}

\paragraph{Results and Discussion}
Figure~\ref{fig:spherecom_all} shows the transition of the best evaluation on \textsc{SphereCOM}.
The CASMOPOLITAN is superior at low budgets, whereas CatCMA generally outperforms the other methods.
We can see that the adaptation mechanisms in CatCMA are more effective as the dimensionality increases.
Figure~\ref{fig:mcproximity_all} shows the result of \textsc{RosenbrockCLO}.
The behavior after the evaluation value reaches $1$ shows that the adaptation mechanisms in CatCMA are particularly effective in the continuous variable optimization.

Figure~\ref{fig:mcproximity_all} shows the result of \textsc{MCProximity}.
In the optimization of \textsc{MCProximity}, the continuous variables $\x$ tend to approach $\boldsymbol{z}$ computed from the categorical variables. Therefore, the convergence of the continuous variables before optimizing the categorical variables leads to performance degradation.
CatCMA performs well, whereas other methods stall in improving the evaluation value.
This result supports that CatCMA can efficiently optimize problems that have dependencies between continuous and categorical variables.

\begin{figure*}[t]
    \centering
    \includegraphics[width=0.95\linewidth]{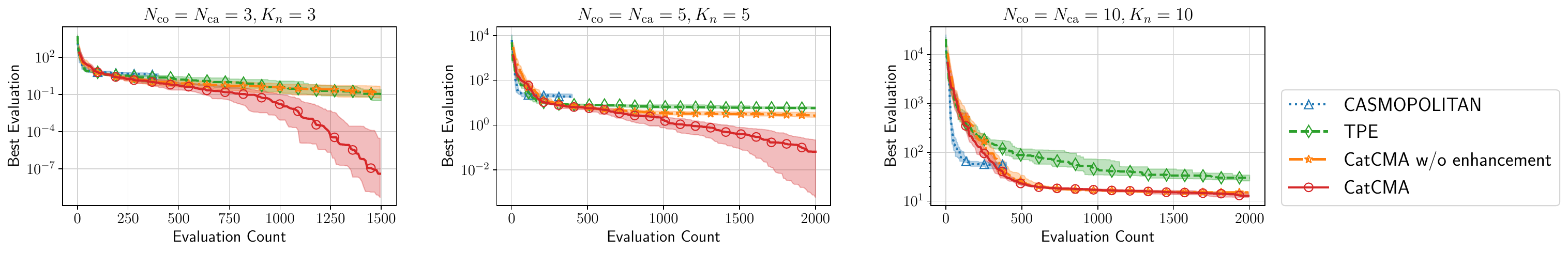}
    \caption{Transition of the best evaluation value on \textsc{RosenbrockCLO}. The line and shaded area denote the medians and interquartile ranges over 20 independent trials, respectively.}
    \label{fig:rosenbrockclo_all}
\end{figure*}

%% file: body/06_conclusion.tex
\section{Conclusion} \label{sec:conclusion}
We proposed CatCMA, a stochastic optimization method for the MC-BBO. The basic update rule of CatCMA is derived from the IGO framework with the joint probability distribution of multivariate Gaussian and categorical distributions.
To balance and accelerate the updating of the joint probability distribution, CatCMA incorporates rank-one update, step-size adaptation, and learning rate adaptation.
While the margin is introduced into the categorical distribution to prevent its premature convergence, we observed empirically that inappropriate margins degrade the performance of CatCMA.
Based on this observation, we analytically derived a promising margin setting and confirmed its effectiveness through numerical experiments.
Numerical experiments also showed that the performance of CatCMA was superior and more robust to problem dimensions compared with the state-of-the-art Bayesian optimization algorithms for the MC-BBO.

CatCMA can introduce the discretization of some continuous variables and the margin correction proposed in \cite{CMAwM:2022} for the multivariate Gaussian distribution, allowing the simultaneous optimization of continuous, integer, and categorical variables.
Although we can easily implement such a combined approach, its hyperparameter setting for efficient optimization is not clear.
In future work, the combined approach requires a comprehensive investigation with different proportions of each variable type, and its application to real-world problems is also important.

%% file: body/99_appendix.tex
\section{Proof of Proposition~\ref{prop:promising_margin}} \label{sec:apdx_proof}
\begin{proof}
When the parameter of the categorical distribution and the margin satisfy \eqref{eq:assum_cat} and \eqref{eq:assum_margin}, the probability that at least one non-optimal category is included in a sample is calculated as follows.
\begin{align*}
    1 - \prod_{n=1}^{\Nca} \qt[t]_{n,1} &= 1 - \prod_{n=1}^{\Nca} \left( 1 - \frac{1}{K_n - 1} \left( 1 - (1 - \xi)^{\frac{1}{\Nca}} \right) (K_n - 1) \right) \\
    &= 1 - \prod_{n=1}^{\Nca} (1-\xi)^{\frac{1}{\Nca}} = \xi
\end{align*}
Then, the random variable $\lambda_\mathrm{non}$ follows the binomial distribution $\mathrm{Bin}(\lambda, \xi)$. The tail probability that $\lambda_\mathrm{non}$ is less than $\lambda - \lfloor \lambda/2 \rfloor$ is computed as
\begin{align}
    \Pr\left(\lambda_\mathrm{non} \leq \lambda - \left\lfloor \frac{\lambda}{2} \right\rfloor \right) = \sum_{i=0}^{\lambda - \left\lfloor \frac{\lambda}{2} \right\rfloor} \binom{\lambda}{i} \xi^i (1 - \xi)^{\lambda - i} \enspace. \label{eq:tail_prob}
\end{align}
\new{
We note that the number of successes in $\lambda'$ Bernoulli trials with success probability $\xi$ follows the binomial distribution $\mathrm{Bin} (\lambda',\xi)$.
We consider independent $\lambda + 2$ Bernoulli trials $b_1,\cdots,b_{\lambda+2} \in \{0, 1\}$ and introduce $X_{k,\xi} = b_1 + \cdots + b_k$.
We consider dependencies between three variables $X_{\lambda,\xi}$, $X_{\lambda+1,\xi}$, and $X_{\lambda+2,\xi}$ in the following calculations.
}
We prove the ineqality $\Pr(X_{\lambda,\xi} \leq \ell) \leq \Pr(X_{\lambda+2, \xi} \leq \ell+1)$ \new{for $\ell \leq \lambda - \lfloor \lambda/2 \rfloor$} and show that \eqref{eq:tail_prob} takes greater than $0.95$ for $\lambda \geq 6$ and $\xi=0.27$.
First, $\Pr(X_{\lambda+1, \xi} \leq \ell)$ can be calculated with $\Pr(X_{\lambda+1, \xi} \leq \ell-1)$ or $\Pr(X_{\lambda, \xi} \leq \ell)$ as follows:
\begin{align}
    &\Pr(X_{\lambda+1, \xi} \leq \ell) \notag \\
    \begin{split}
        &= \Pr(X_{\lambda+1, \xi} \leq \ell \mid X_{\lambda, \xi} \leq \ell-1) \Pr(X_{\lambda, \xi} \leq \ell-1) \\
        &\qquad + \Pr(X_{\lambda+1, \xi} \leq \ell \mid X_{\lambda, \xi} = \ell-1) \Pr(X_{\lambda, \xi} = \ell-1) \qquad
    \end{split} \\
    &= \Pr(X_{\lambda, \xi} \leq \ell-1) + (1-\xi) \binom{\lambda}{\ell} \xi^{\ell} (1-\xi)^{\lambda-\ell} \label{eq:binom_1_1} \\
    &= \Pr(X_{\lambda, \xi} \leq \ell) - \binom{\lambda}{\ell} \xi^{\ell} (1-\xi)^{\lambda-\ell} + (1-\xi) \binom{\lambda}{\ell} \xi^{\ell} (1-\xi)^{n-\ell} \\
    &= \Pr(X_{\lambda, \xi} \leq \ell) - \binom{\lambda}{\ell} \xi^{\ell+1} (1-\xi)^{\lambda-\ell} \label{eq:binom_1_0}
\end{align}
Then, $\Pr(X_{\lambda+2, \xi} \leq \ell+1)$ can be computed as follows.
\begin{align}
    &\Pr(X_{\lambda+2, \xi} \leq \ell+1) \label{eq:binom_first} \\
    &= \Pr(X_{\lambda+1, \xi} \leq \ell) + (1-\xi) \binom{\lambda+1}{\ell+1} \xi^{\ell+1} (1-\xi)^{\lambda-\ell} \label{eq:binom_second} \\
    \begin{split}
        &= \Pr(X_{\lambda, \xi} \leq \ell) - \binom{\lambda}{\ell} \xi^{\ell+1} (1-\xi)^{\lambda-\ell} \\
        &\qquad\qquad\qquad\qquad + (1-\xi) \binom{\lambda+1}{\ell+1} \xi^{\ell+1} (1-\xi)^{\lambda-\ell}
    \end{split} \label{eq:binom_third} \\
    &= \Pr(X_{\lambda, \xi} \leq \ell) + \left( \frac{(1-\xi)(\lambda+1)}{\ell+1} - 1 \right) \binom{\lambda}{\ell} \xi^{\ell+1} (1-\xi)^{\lambda-\ell} \label{eq:binom_npk}
\end{align}
We note that the transformation of \eqref{eq:binom_first} to \eqref{eq:binom_second} uses \eqref{eq:binom_1_1}, and the transformation of \eqref{eq:binom_second} to \eqref{eq:binom_third} uses \eqref{eq:binom_1_0}.
If $\lambda \geq 6$, $\ell \leq \lambda - \lfloor \lambda/2 \rfloor$, and $\xi = 0.27$, the second term of \eqref{eq:binom_npk} is positive. Then, $\Pr(X_{\lambda,\xi} \leq k) \leq \Pr(X_{\lambda+2,\xi} \leq k+1)$ holds and we obtain the following tail probabilities. 
\begin{align}
    0.9508\ldots = \Pr(X_{6,0.27} \leq 3) &\leq \Pr(X_{8,0.27} \leq 4) \leq \cdots \\
    0.9818\ldots = \Pr(X_{7,0.27} \leq 4) &\leq \Pr(X_{9,0.27} \leq 5) \leq \cdots
\end{align}
Thus, $\Pr(\lambda_\mathrm{non} \leq \lambda - \lfloor \lambda/2 \rfloor) \geq 0.95$ holds.
\end{proof}

\section{Additional Experimental Results}
\begin{figure*}[b]
    \centering
    \includegraphics[width=0.95\linewidth]{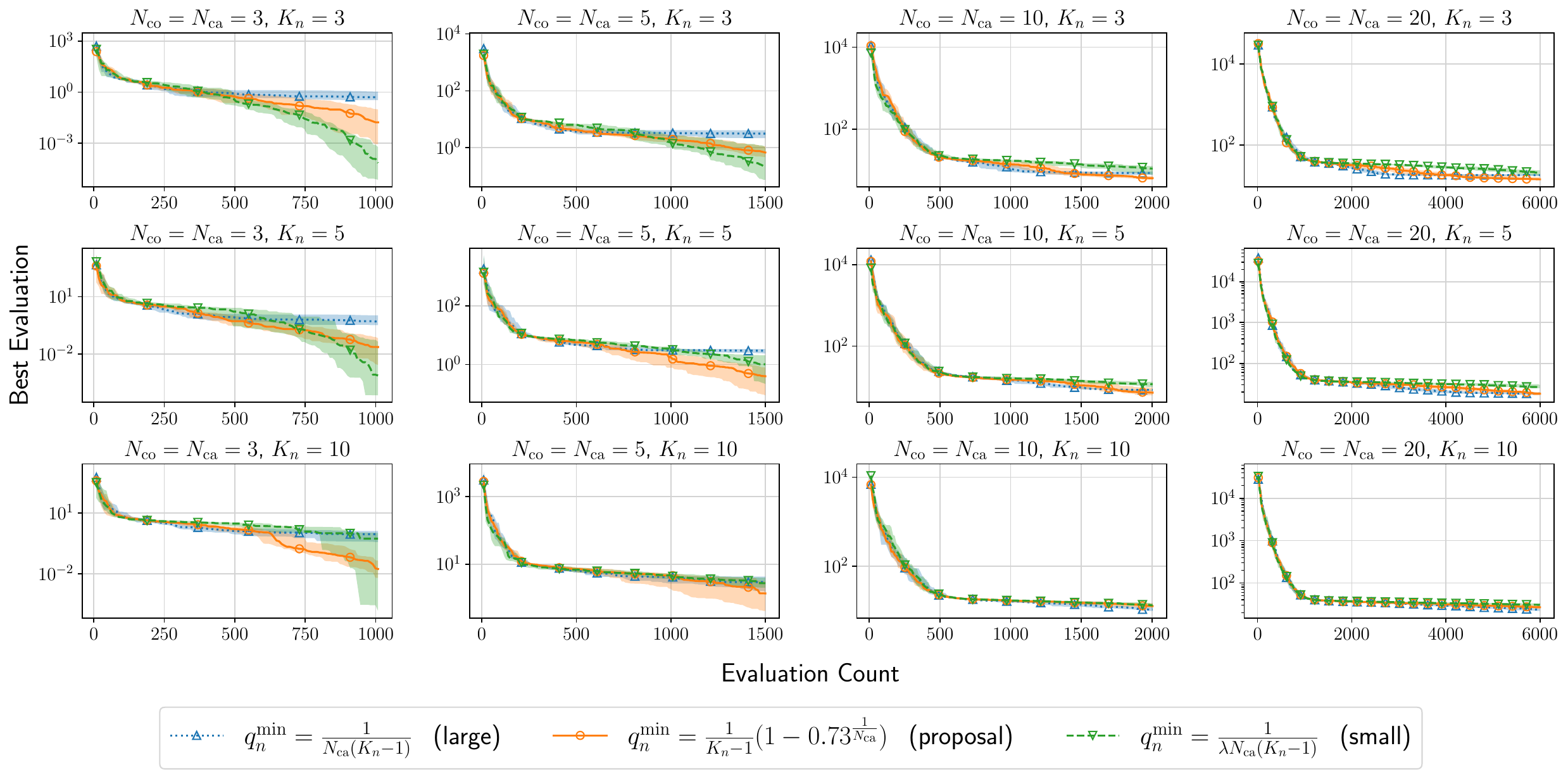}
    \caption{The transition of the best evaluation value on \textsc{RosenbrockCLO}. The line and shaded area denote the medians and interquartile ranges over 20 independent trials, respectively. The experimental setting is shown in Section~\ref{ssec:exp_margin}.}
    \label{fig:rosen_grid}
\end{figure*}

\begin{figure*}[b]
    \centering
    \includegraphics[width=0.95\linewidth]{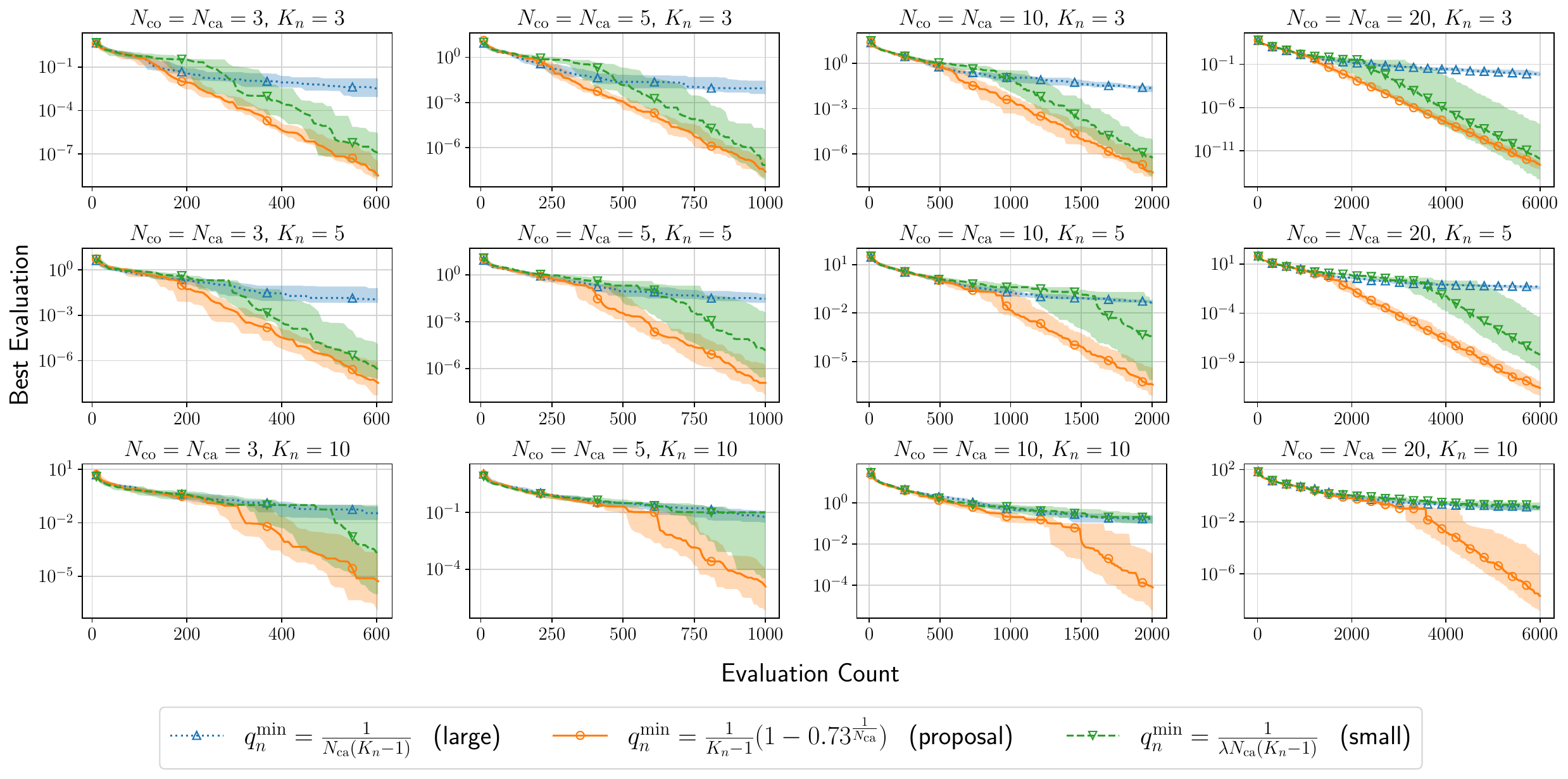}
    \caption{The transition of the best evaluation value on \textsc{MCProximity}. The line and shaded area denote the medians and interquartile ranges over 20 independent trials, respectively. The experimental setting is shown in Section~\ref{ssec:exp_margin}.}
    \label{fig:mcproximity_grid}
\end{figure*}

\new{
We introduce the comparison experiments between small, large, and recommended margin settings on \textsc{RosenbrockCLO} and \textsc{MCProximity} that are relatively complex problems compared with \textsc{SphereCOM}.
The experimental setting is the same as for \textsc{SphereCOM} in Section~\ref{ssec:exp_margin}.
Figure~\ref{fig:rosen_grid} and \ref{fig:mcproximity_grid} show the the experimental results on \textsc{RosenbrockCLO} and \textsc{MCProximity}.
Form the experimental result, we can confirm the effectiveness of recommended margin setting.
}


